%% file: neurips_2019.tex
\documentclass{article}

\usepackage[final,nonatbib]{neurips_2019}

\usepackage[utf8]{inputenc} 
\usepackage[T1]{fontenc}    
\usepackage{hyperref}       
\usepackage{url}            
\usepackage{booktabs}       
\usepackage{amsfonts}       
\usepackage{nicefrac}       
\usepackage{microtype}      
\usepackage{algorithm}
\usepackage{algorithmicx}
\usepackage{algpseudocode}

\usepackage{subfig}
\usepackage{varwidth}
\usepackage{capt-of}
\usepackage{xspace}
\usepackage{caption}
\usepackage{graphicx}
\usepackage{mathtools}
\usepackage{pgfplots}
\usepackage{marvosym,listings,etoolbox}

\newcommand{\alg}{{\textsc{DynMDND}}\xspace}

\usepackage{amsthm,amssymb,amsmath,mathtools}

\usepackage{authblk}
\title{Sequential Edge Clustering in Temporal Multigraphs}

\author[1]{\small Elahe Ghalebi}
\author[2]{\small Hamidreza Mahyar}
\author[1]{\small Radu Grosu}
\author[3,4]{\small Graham W. Taylor}
\author[5]{\small Sinead A. Williamson}

\affil[1]{\footnotesize Technical University of Vienna, Austria}
\affil[2]{\footnotesize Boston University, USA}
\affil[3]{\footnotesize University of Guelph, Canada}
\affil[4]{\footnotesize Vector Institute for Artificial Intelligence, Canada}
\affil[5]{\footnotesize University of Texas, Austin, USA}

\begin{document}
\maketitle
\begin{abstract}
Interaction graphs, such as those recording emails between individuals or transactions between institutions, tend to be sparse yet structured, and often grow in an unbounded manner. Such behavior can be well-captured by structured, nonparametric edge-exchangeable graphs. However, such exchangeable models necessarily ignore temporal dynamics in the network. We propose a dynamic nonparametric model for interaction graphs that combines properties of edge-exchangeable models with dynamic clustering patterns that tend to reinforce recent behavioral patterns. We show that our method yields improved held-out likelihood over stationary variants, and impressive predictive performance against a range of state-of-the-art dynamic interaction graph models. 
\end{abstract}
\section{Introduction}\label{sec:intro}
\input{intro.tex}
\section{Related Work}\label{sec:related}
\input{related.tex}
\section{Sequential Edge Clustering in Temporal Multigraphs}\label{sec:model}
\input{model.tex}
\section{Experiments}\label{sec:experiments}
\input{experiments.tex}
\section{Conclusion}
We have presented a new model for interaction networks that can be represented in terms of sequences of links, such as email interaction graphs and collaboration graphs. Using a nonparametric sequence of links makes our model well-suited to predicting future links, and unlike many vertex-based graph models allows for an unbounded number of vertices.
\bibliographystyle{unsrt}
\bibliography{neurips_2019}

\end{document}

%% file: intro.tex
Many forms of social interaction can be represented in terms of a multigraph, where each individual interaction corresponds to an edge in the graph, and repeated interactions may occur between two individuals. For example, we might have multigraphs where the value of an edge corresponds to the number of emails between two individuals, or the number of packets sent between two hosts. 
\\
Traditional clustering algorithms such as K-means, hierarchical clustering, and stochastic block models have been successfully applied to cluster the nodes of a finite static graph, but only a few have considered edge clustering. Recently, the class of edge exchangeable graphs \cite{CaiCampbellBroderick2016,CraneDempsey2018,Williamson2016} have been proposed for modeling networks as exchangeable sequences of edges. These models are able to capture many properties of large-scale social networks, such as sparsity, community structure, and power-law degree distribution. 
\\
Being explicit models for sequences of edges, edge-exchangeable models are appropriate for networks that grow over time: we can add more edges by expanding the sequence, and their nonparametric nature means that we expect to introduce previously unseen vertices as the network expands. However, their exchangeable nature precludes graphs whose properties change over time. In practice, the dynamics of social interactions tend to vary over time. In particular, in models that aim to capture community dynamics, the popularity of a given community can wax and wane over time.
\\
We propose a new model for multigraphs with clustered edges, that breaks the exchangeability of existing models by preferentially assigning edges to clusters that have been recently active. We show that incorporating dynamics using a mechanism based on the distance-dependent Chinese restaurant process (ddCRP) \cite{BleiFrazier2011} leads to improved test-set predictive likelihood over exchangeable models. Further, when used in a link prediction task, we show improved performance over both its exchangeable counterpart and a range of state-of-the-art dynamic network models.
\\
We begin in Section~\ref{sec:intro} and \ref{sec:related} by providing a general introduction to Bayesian models for multigraphs, in both the static and dynamic setting, plus a review of related works. We then introduce our model in Section~\ref{sec:model}, before empirically evaluating our approach in Section~\ref{sec:experiments}.

%% file: related.tex
\textbf{Bayesian Models for Multigraphs.} Most Bayesian models for graphs and multigraphs fall under the vertex-exchangeable framework, where the distribution over the adjacency matrix is invariant to jointly permuting the row and column indices. In this class, we have models such as the stochastic block model, where vertices are clustered into finitely many communities and a parameter is associated with each community-community pair \cite{Snijders:Nowicki:1997,Karrer:Newman:2011}; the infinite relational model, where the number of communities is unbounded \cite{irm}; mixed-membership stochastic block models, where edges are generated according to an admixture model \cite{airoldi2008mixed}; the latent feature relational model, where parameter are distributed according to a latent feature model based on the Indian buffet process \cite{miller2009nonparametric}; and Poisson factor analysis, where
the parameter are distributed according to a gamma process-based latent factor model \cite{ZhouCarin2013,gopalan2015scalable}. While these models are able to capture interesting community structure, the resulting graphs are dense or empty almost surely \cite{Aldous1981, Hoover:1979}. This makes them a poor choice for large real-world networks, which are typically sparse. Further, since they explicitly model zero-edges, they are not well-suited to link prediction tasks.
\\
An alternative way of constructing multigraphs is using an exchangeable sequence of edges \cite{CaiCampbellBroderick2016,CraneDempsey2018,Williamson2016}. Here, we assume the edges are generated by sequentially sampling pairs of vertices. These pairs are \textit{iid} given some nonparametric prior, such as a Dirichlet process, a normalized generalized gamma process, or a Pitman Yor process. This construction means multigraphs can grow over time by adding new edges, making them well-suited for link prediction tasks. Appropriate choices of priors can yield sparsity and power law degree distribution. 
\\
\textbf{Models for Dynamic Graphs.}
There has been significant research attention on dynamic (time-evolving) network modelling, ranging from non-Bayesian methods such as dynamic extensions of the exponential random graph model (ERGM)~\cite{guo2007recovering}, or matrix and tensor factorization-based methods~\cite{dunlavy2011temporal}, to Bayesian latent variable models~\cite{pmlr-v2-sarkar07a,ishiguro2010dynamic,sarkar2014nonparametric,durante2014nonparametric,schein2016bayesian, PallaCaronTeh2016,ng2017dynamic,yang2018dependent}. A common approach relies on the extensions of \emph{static} network models to a \emph{dynamic} framework \cite{ghalebi2018dynamic}. We focus here on dynamic extensions of Bayesian models of the forms discussed above.
\\
Most dynamic Bayesian networks extend jointly vertex-exchangeable graphs. For example, \cite{xu2014dynamic} extends the stochastic blockmodel using an extended Kalman filter (EFK) based algorithm, and the stochastic block transition model \cite{xu2015sbtm} relaxes a hidden Markov assumption on the edge-level dynamics, allowing the presence or absence of edges to directly influence future edge probabilities. Several methods have also been used to incorporate temporal dynamics into the mixed membership stochastic blockmodel framework \cite{fu2009dynamic,xing2010state,ho2011evolving} and the latent feature relational model \cite{foulds2011dynamic,heaukulani2013dynamic,kim2013nonparametric}. Lately, several models have extended Poisson factor analysis. The dynamic gamma process Poisson factorization (DGPPF)~\cite{acharya2015nonparametric} introduces dependency by incorporating a Markov chain of marginally gamma random variables into the latent representation. The dynamic Poisson gamma model (DPGM)~\cite{yang2018poisson} extends a bilinear form of Poisson factor analysis~\cite{zhou2015infinite} in a similar manner. The dynamic relational gamma process model (DRGPM)~\cite{yang2018dependent} also incorporates a temporally dependent thinning process. 

%% file: model.tex
Our model extends the Mixture of Dirichlet Network Distributions (MDND) \cite{Williamson2016}, an edge-exchangeable model that clusters edges to create community-like structure.  The MDND is based on a sequence of Dirichlet processes: one controls the global popularity of vertices; one controls a distribution over cluster indicators; and two per cluster control the cluster-specific distributions over ``sender'' and ``'recipient'' vertices. Any of these distributions could be replaced with dynamic or dependent clustering models to generate a temporally evolving graph. In practice, replacing all of the distributions with dynamic alternatives is likely to lead to overspecification of the dependencies, making inference challenging.
\\
We choose to retain stationary models for $H$, $A_k$ and $B_k$, implying that a cluster's representation stays stable over time, and allow the cluster popularities to vary by making the sequence $z_1, z_2, \dots$ time-varying.  A number of methods exist for incorporating temporal dynamics into the Dirichlet process or the related Chinese restaurant process (CRP), \textit{e.g.} \cite{maceachern2000dependent,lin2010construction,Ren:2008:DHD:1390156.1390260}.  For our purposes, we choose to use the distance-dependent CRP (ddCRP) \cite{BleiFrazier2011}, as it captures the property that we are likely to see clusters that have appeared recently. Recall that the CRP can be described in terms of a restaurant analogy, where customers select tables (clusters) proportional to the number of people already seated at that table, or sit at a new table with probability proportional to a concentration parameter $\alpha$. The ddCRP modifies this by encouraging customers to sit next to ``similar'' customers. In a time-dependent setting, similarity is evaluated based on arrival time using some non-negative, non-increasing decay function $f$ such that $f(\infty)=0$. Concretely, given $t_i,t_j$ arrival times of $i$ and $j$,
\begin{equation*}
d_{i, j} = \begin{cases} t_i - t_j & t_i \geq t_j\\ \infty &
 \mbox{otherwise}\end{cases}
\end{equation*}
\\
Then the $i$-th customer picks a customer $c_i$ to sit next to (and therefore a cluster) according to
\begin{equation}
P(c_i = j) \propto \begin{cases} f(d_{i,j}) & i\neq j\\ \alpha & i=j\end{cases}
\end{equation}\label{eq:seating}
\\
In an interaction network context, this implies that we are likely to see modes of communication that have been popular in recent time periods, over modes of communication that have fallen out of popularity. Another reason to favor the ddCRP is ease of inference: its construction lends itself to an easy-to-implement Gibbs sampler, allowing us to apply our method to larger graphs. By contrast, many other dependent Dirichlet processes have much more complicated inference algorithms, which would limit scalability. While the ddCRP does not yield marginal invariance (i.e.\ it is not invariant to adding edges at previously observed time points), this is not a concern in our setting, since we are typically able to observe past instances of the full graph, and are interested in predicting future edges. 
\\
Incorporating the ddCRP into the MDND yields the following generative process:

\begin{equation}\label{eqn:mdnd}
\begin{aligned}[c]
	H & \coloneqq \sum_{i=1}^\infty h_i\delta_{\theta_i} &\sim& ~\text{DP}(\gamma,\Theta)  \\
	A_k & \coloneqq \sum_{i=1}^\infty a_{i}\delta_{\theta_i} &\sim& ~\text{DP}(\tau, H) \\
	B_k & \coloneqq \sum_{i=1}^\infty b_{i}\delta_{\theta_i} &\sim& ~\text{DP}(\tau,H) 
\end{aligned}
\qquad
\qquad
\begin{aligned}[c]
    P(c_i = j) \propto& \begin{cases} f(d_{i,j}) & i\neq j\\ \alpha & i=j \end{cases}\\
    z_i =& \begin{cases} z_{c_i} & c_i\neq i\\ i & c_i=i\end{cases}\\
	s_n \sim& A_{z_n}\\
	r_n \sim& B_{z_n}
\end{aligned}
\vspace{-2pt}
\end{equation}

We perform inference using an MCMC scheme based on that developed in \cite{BleiFrazier2011}.

%% file: experiments.tex
\begin{figure}
    \vspace{-.3cm}
	\centering  
	\subfloat[MAP@k\label{subfig:maps}]{\includegraphics[scale=0.2,trim= 5mm 10mm 0mm 0mm] {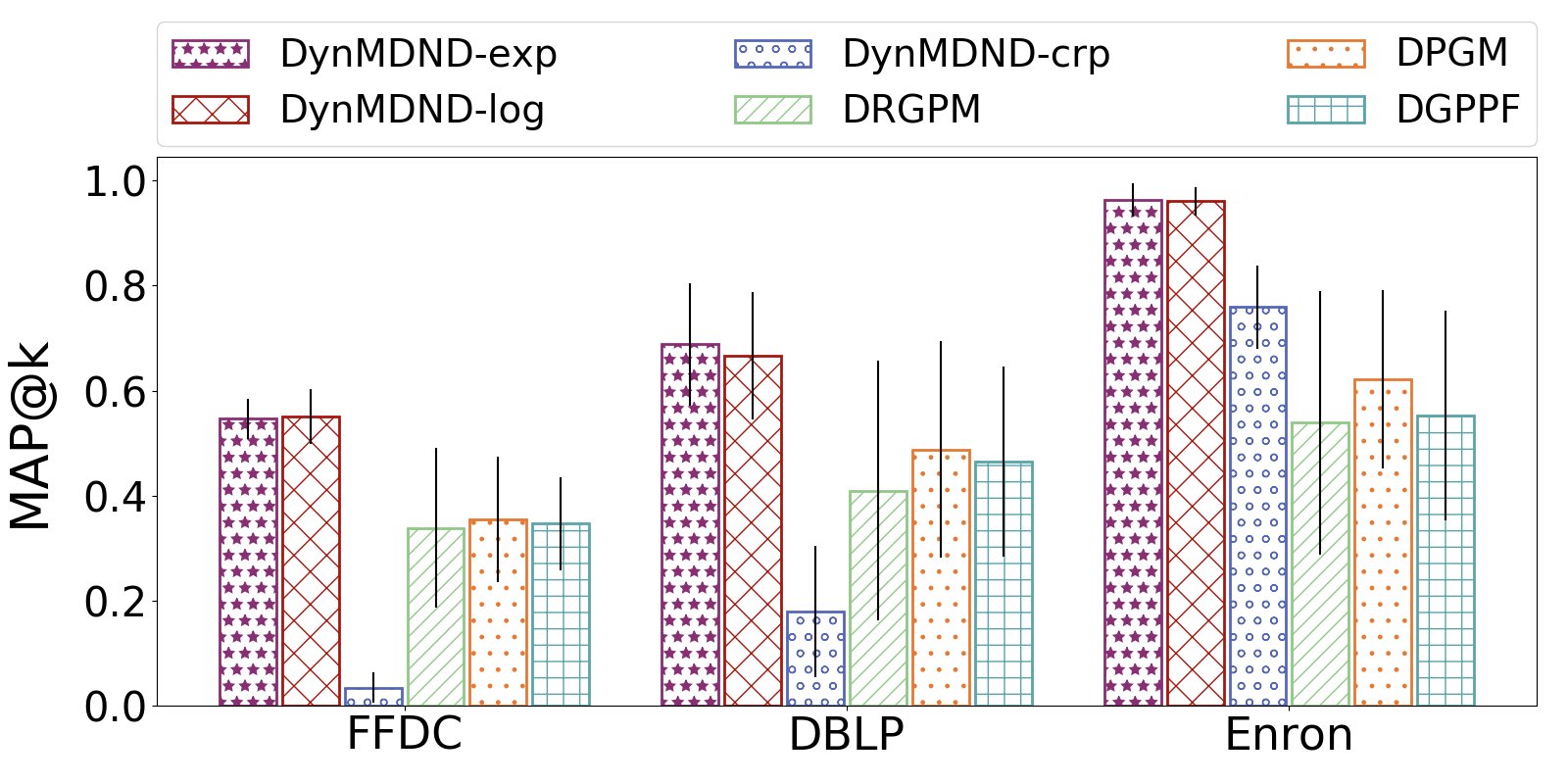}}
	\subfloat[Hits@k\label{subfig:maps}]{\includegraphics[scale=.2,trim= 5mm 10mm 0mm 0mm] {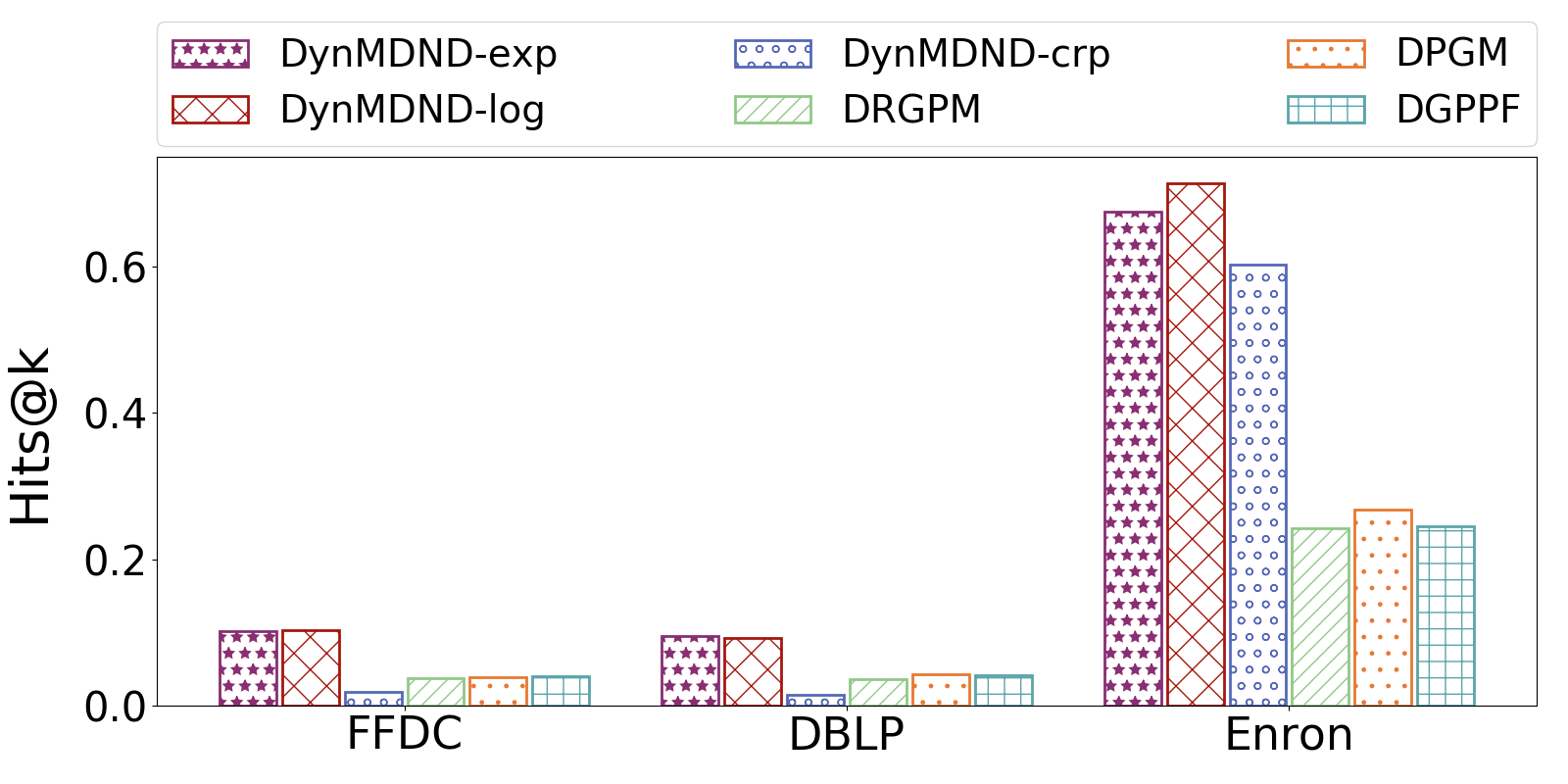}}\\
	\subfloat[F1 Score\label{subfig:maps}]{\includegraphics[scale=.2,trim= 5mm 10mm 0mm 0mm] {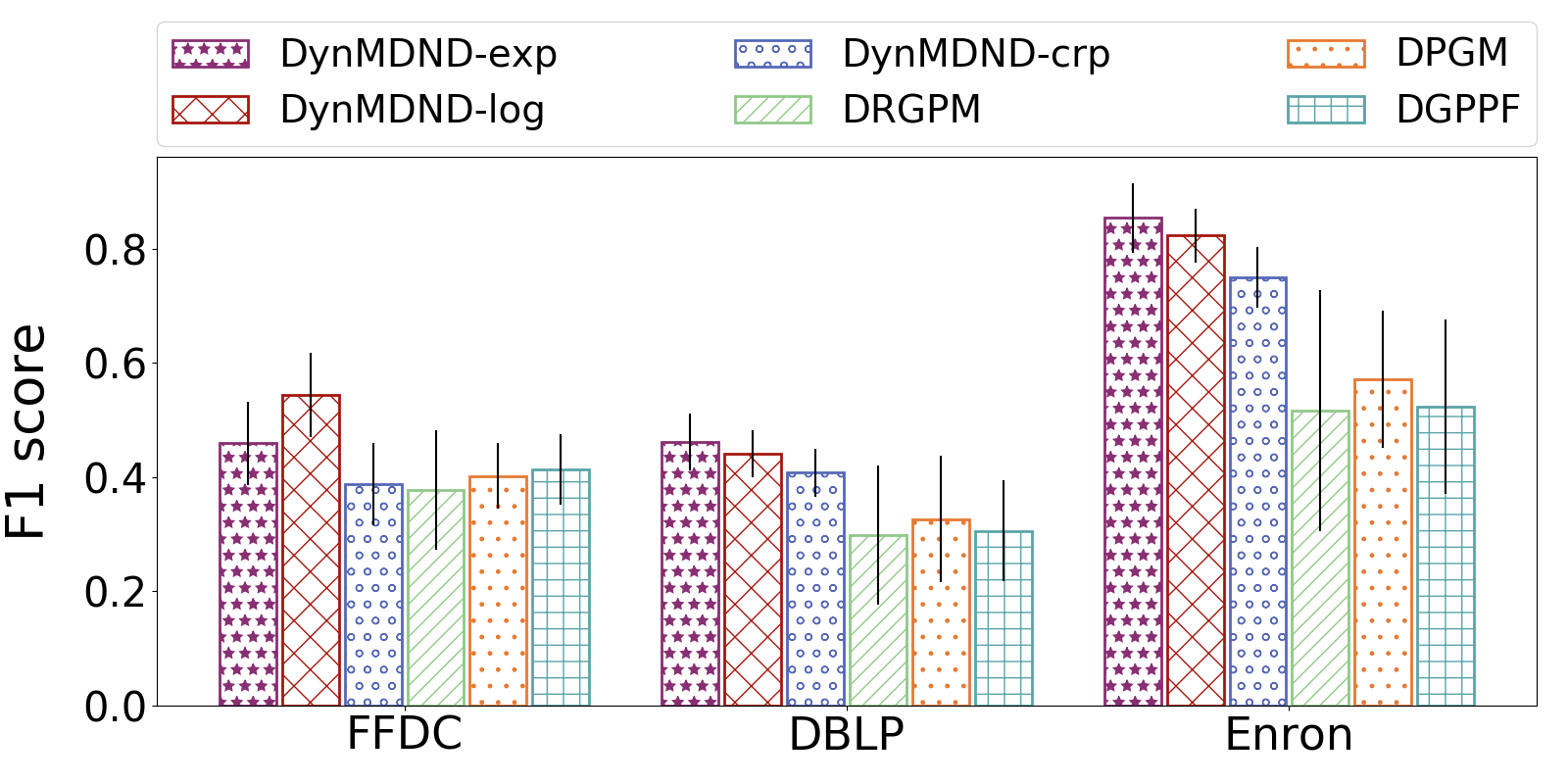}}
	\caption{Performance of \alg for dynamic link prediction compared to \textsc{DRPGM}, \textsc{DPGM} and \textsc{DGPPF} on the three datasets. }
	\label{fig:hitmap}
	\vspace{-.5cm}
\end{figure}

\textbf{Datasets.} We evaluate our model on four real-world networks: (1) Face-to-Face Dynamic Contacts (FFDC)\footnote{\url{http://www.sociopatterns.org/datasets/high-school-contact-and-friendship-networks/}} \cite{mastrandrea2015contact}; $M = 180$ students with $N = 8,332$ communications for $t = 7$ school days. (2) Social Evolution (SocialEv)\footnote{\url{http://realitycommons.media.mit.edu/socialevolution.html}} \cite{madan2011sensing}; $M = 68$ nodes and $N = 399$ links. This network has a high clustering coefficient and about $2\times10e6$ events over $t = 6$ time slots.
(3) DBLP \cite{asur2009event}; $M = 324$ most connected authors over all time periods which contains $N = 11,154$ edges.
(4) Enron\footnote{\url{https://www.cs.cmu.edu/~enron/}}; $N = 8,534$ interactions among $M = 151$ users over 38 months (May 1999- June 2002).
\\
\textbf{Metrics.} We use F1 score, Map@$k$ and Hit@$k$. F1 score is 2$\times$(precision$\times$recall)/(precision+recall). Precision is the fraction of edges in the future network present in the true network, Recall is the fraction of edges of the true network present in the future network. \!MAP@$k$ is the classical mean average precision measure and Hits@$k$ is the rate of the top-$k$ ranked edges.
\\
\textbf{Comparison.} In addition to the exchangeable MDND, we consider 
three state-of-the-art network models, discussed in Section~\ref{sec:related}: DRGPM \cite{yang2018dependent}, DPGM \cite{yang2018poisson}, and DGPPF \cite{acharya2015nonparametric}. None of these models are designed for explicit link prediction, yet can be modified to give predictions using the above procedure of selecting the $N$ highest probability edges. These models also have the limitation of assuming a fixed number of vertices. 
\\
\textbf{Results.} Figure \ref{fig:hitmap} illustrates the F1 score, Map@$k$ and Hits@$k$ for the proposed model, \alg, with all three decay types, \textsc{Exponential}, \textsc{Logistic} and \textsc{CRP} compared to \textsc{DRPGM}, \textsc{DPGM} and \textsc{DGPPF} for dynamic link prediction. We use the networks of time slots 1 to $T$ as a training set and predict the network edges of time slot $T+1$. We report the results on the three datasets: FFDC, DBLP and Enron, using time interval of one day, one year and one month, respectively. For each task, we repeat the experiments 10 times and report the mean and standard deviation of each evaluation metric.
\\
We see that \alg significantly outperforms \textsc{DRPGM}, \textsc{DPGM} and \textsc{DGPPF} on all metrics, for the task of dynamic link prediction. We hypothesise that this is due to several reasons. First, \alg is explicitly designed in terms of a predictive distribution over edges, making it well-suited to predicting future edges. Second, \alg is able to increase the number of vertices over time, and is likely better able to capture natural network growth. Conversely,  the other methods assume the number of vertices is fixed.
\\
Table \ref{tbl:log_lhood} shows the predictive log likelihood computed by \alg using two different decays (\textit{i.e.} \textsc{Exponential} and \textsc{Logistic}) in comparison with the \textsc{CRP} decay function. The other comparison methods do not incorporate such a log likelihood. At each time slot $T$, we use 80\% of the network data for training the model and the remaining 20\% for the test set. It can be seen that considering time dependency into our mixture models results in a better log likelihood for the task of prediction. We use a sample size of 1000. We repeat the experiments 10 times and report the mean and standard deviation of the results for the four real networks. 

\begin{table}[t]
    \centering
    \caption{Predictive Log Likelihood for the held-out (test set) data computed by \alg with three different decay functions on four real-world networks (mean $\pm$ standard error).}
    \small
    \begin{tabular}[h]{cccc}
		\toprule
		Dataset &\textsc{MDND} & \textsc{\alg-Logistic} & \textsc{\alg-Exponential}\\
		\midrule
		FFDC & -383,094.6$\pm$146 & \textbf{-286,833.9$\pm$220} &-344,683.9$\pm$105\\
		Enron & -1032.94 $\pm$ 147.18 & \textbf{-640.73$\pm$90.51}  & -700.88$\pm$ 22.97 \\
		DBLP & -980,928.4$\pm$532.1 &-798,568.4$\pm$998.5  & \textbf{-649,521$\pm$195.0} \\
		SocialEv & -173,708.1$\pm$223.6 & -23,087.3$\pm$91.0& \textbf{-19,820.1$\pm$93.6}\\
		\bottomrule
 \end{tabular}
    \label{tbl:log_lhood}
\end{table}